\documentclass{article}
\usepackage{ijcai23}

\usepackage{times}
\usepackage{soul}
\usepackage{url}
\usepackage[hidelinks]{hyperref}
\usepackage[utf8]{inputenc}
\usepackage[small]{caption}
\usepackage{graphicx}
\usepackage{amsmath}
\usepackage{amsthm}
\usepackage{booktabs}
\usepackage[switch]{lineno}

\usepackage{graphicx}
\usepackage{amsmath}
\usepackage{amssymb}
\usepackage{booktabs}

\usepackage[utf8]{inputenc} 
\usepackage[T1]{fontenc}    
\usepackage{nicefrac}       
\usepackage{microtype}      
\usepackage{xcolor}         
\usepackage{makecell}

\usepackage{bm}
\usepackage{multirow}
\usepackage{tabu}
\usepackage{siunitx}
\usepackage{adjustbox}
\usepackage{subcaption}
\usepackage{siunitx}
\usepackage{xcolor}
\usepackage{colortbl}
\usepackage{color}
\usepackage{subcaption}
\usepackage{pifont}
\definecolor{Gray}{gray}{0.95}
\definecolor{Cyan}{rgb}{0.88,1,1}

\usepackage{tabu}           
\usepackage{multirow}
\usepackage{booktabs}
\usepackage{makecell}
\usepackage{wrapfig}

\usepackage[ruled,vlined,linesnumbered]{algorithm2e}
\usepackage{listings}
\usepackage{comment}

\title{Adaptive Sparse ViT: Towards Learnable Adaptive Token Pruning by Fully Exploiting Self-Attention}

\author{
Xiangcheng~Liu$^1$\textsuperscript{\thanks{Equal contribution} \thanks{Interns at the Institute of Deep Learning, Baidu Research}}
\and
Tianyi Wu$^2$\textsuperscript{*}
\And
Guodong Guo$^3$\textsuperscript{\thanks{Corresponding author}}
\affiliations
$^1$Peking University\\
$^2$Baidu Autonomous Driving Technology Department (ADT)\\
$^3$Institute of Deep Learning, Baidu Research
\emails
liuxiangcheng@stu.pku.edu.cn,
wutianyi01@baidu.com,
Guodong.Guo@mail.wvu.edu
}

\newcommand\cb[1]{\color{blue} #1}

\DeclareMathOperator{\MHSA}{MHSA}
\DeclareMathOperator{\ascore}{A}
\DeclareMathOperator{\CTX}{Context}
\DeclareMathOperator{\FFN}{FFN}
\DeclareMathOperator{\LN}{LN}

\DeclareMathOperator{\softmax}{Softmax}
\DeclareMathOperator{\sigmoid}{Sigmoid}
\DeclareMathOperator{\flops}{FLOPs}
\DeclareMathOperator{\kld}{KL}
\DeclareMathOperator{\CE}{CrossEntropy}

\begin{document}

\maketitle

\begin{abstract}
    Vision transformer has emerged as a new paradigm in computer vision, showing excellent performance while accompanied by expensive computational cost. Image token pruning is one of the main approaches for ViT compression, due to the facts that the complexity is quadratic with respect to the token number, and many tokens containing only background regions do not truly contribute to the final prediction. Existing works either rely on additional modules to score the importance of individual tokens, or implement a fixed ratio pruning strategy for different input instances. In this work, we propose an adaptive sparse token pruning framework with a minimal cost. Specifically, we firstly propose an inexpensive attention head importance weighted class attention scoring mechanism. Then, learnable parameters are inserted as thresholds to distinguish informative tokens from unimportant ones. By comparing token attention scores and thresholds, we can discard useless tokens hierarchically and thus accelerate inference. The learnable thresholds are optimized in budget-aware training to balance accuracy and complexity, performing the corresponding pruning configurations for different input instances. Extensive experiments demonstrate the effectiveness of our approach. Our method improves the throughput of DeiT-S by 50\% and brings only 0.2\% drop in top-1 accuracy, which achieves a better trade-off between accuracy and latency than the previous methods.
\end{abstract}

\section{Introduction}

Recently, Vision Transformer (ViT) has made remarkable progress on image classification \cite{dosovitskiy2020image,touvron2021training,liu2021swin}, object detection \cite{carion2020end,zhu2020deformable}, semantic segmentation \cite{SETR,xie2021segformer}, and other vision tasks. 
However, as the model complexity is quadratic to the number of tokens, ViT suffers from expensive computational costs, which limits its application and deployment.

\begin{figure}[t]
\centering
\includegraphics[width=1.0\columnwidth]{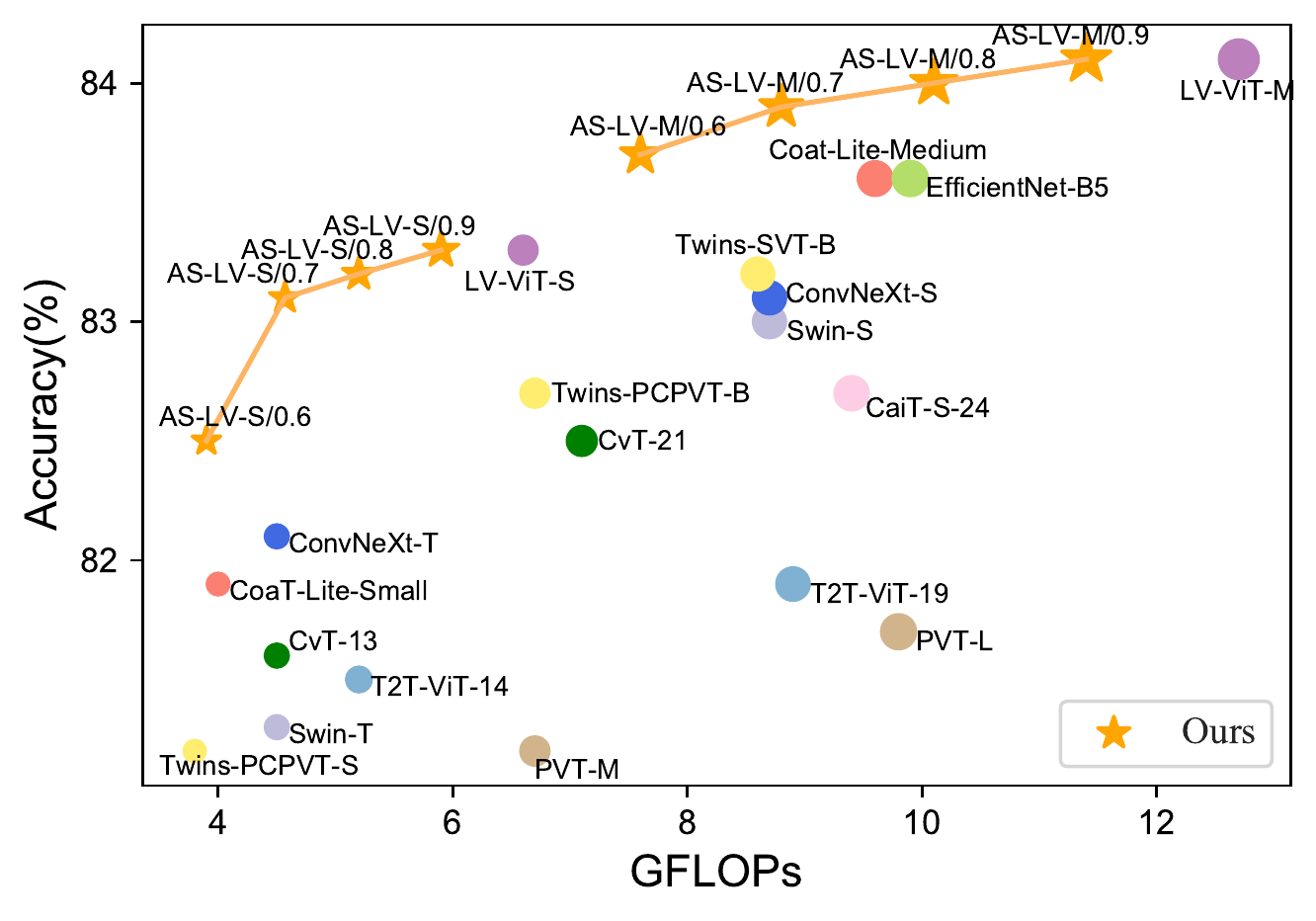}
\caption{Trade-offs between complexity and top-1 accuracy for different models on ImageNet.}
\label{fig:sota}
\end{figure}

\begin{figure}[t]
\centering
\includegraphics[width=0.96\columnwidth]{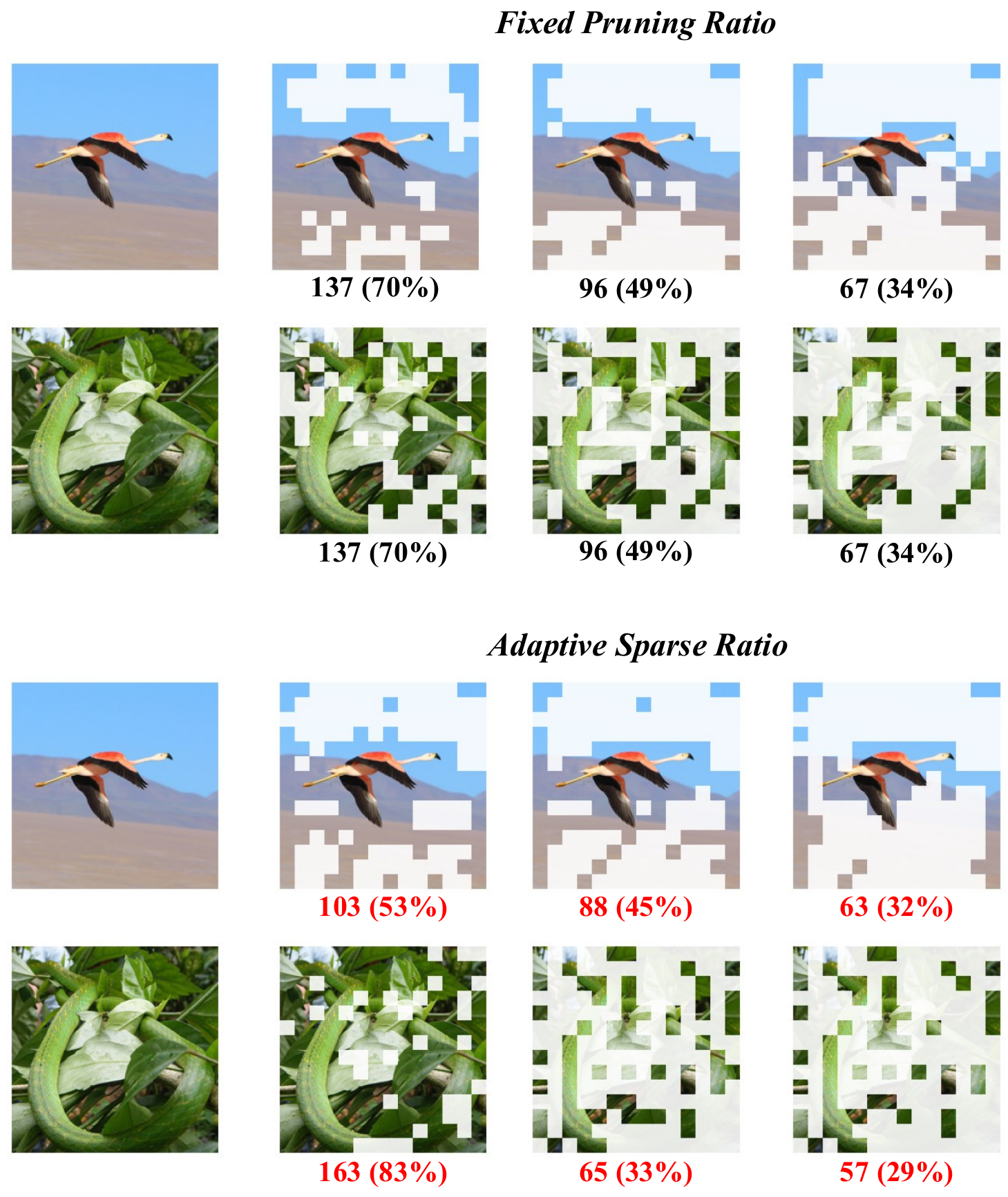}
\caption{Comparison of fixed pruning rate (up) and adaptive sparse rate (down). The number denotes the amount and percentage of tokens kept in the current stage.}
\label{fig:compare}
\end{figure}

Not all image patches are helpful for the final prediction. For instance, the large number of image tokens in the background region do not contribute to the recognition and can be pruned during the inference process, greatly accelerating the model runtime without significant impact on performance. Token pruning has attracted the interests of many researchers. We classify the existing methods according to whether additional calculations are introduced to evaluate the token score. 
Evo-ViT \cite{xu2022evovit} utilizes class attention~\cite{xu2022evovit} to estimate token score and develop a novel slow-fast token evolution approach to improve the throughput of ViT. EViT \cite{liang2022evit} employ a similar method to measure token importance and fuse discarded tokens. Both of these approaches require manually specifying the pruning ratio for each stage, and perform the same pruning policy for different input instances, which may result in simple samples being under-pruned or complex samples being over-pruned in the beginning stages, as illustrated in Figure \ref{fig:compare} top. Another type of work identifies token importance via extra measures. DynamicViT \cite{rao2021dynamicvit} prune tokens in a fixed ratio by inserting lightweight predictors to predict token scores. IA-RED$^2$ \cite{pan2021ia} introduce a multi-head interpreter and employed reinforcement learning to generate pruning scheme for each token. A-ViT~\cite{yin2021adavit} compute halting scores for all tokens to adaptively discard unimportant tokens. The latter two achieve sample-adaptive pruning at the cost of additional computation of significance scores for all tokens.

In this work, we propose an adaptive token sparse framework for ViT acceleration, named AS-ViT, which fully exploits Multi-Head Self-Attention ($\MHSA$) to estimate token importance scores, and uses a minimal cost, only three learnable thresholds, to perform the corresponding token pruning policy for a specific input instance as shown in Figure \ref{fig:compare} bottom. Specifically, we first propose the attention head importance weighted class attention score. It uses the intermediate results of $\MHSA$ to calculate token-level head importance, which is then multiplied as a weighting factor on the class attention~\cite{xu2022evovit} score to better identify informative tokens. Then, we insert three learnable thresholds in ViT hierarchically. Only tokens with a score greater than the current threshold will be kept, and the pruned token will not participate in the later computations. Lastly, we propose a budget-aware loss to optimize the thresholds to achieve a trade-off between accuracy and computational effort across the dataset. During testing, the learnable threshold is fixed and we just compare the threshold and individual token scores to actually discard uninformative tokens for hardware acceleration. We investigate the intrinsic features of $\MHSA$ and reuse its computational results to evaluate the informativeness of the token. Threshold-based comparison avoids sorting all tokens. We use the above method to minimize the cost of token pruning.

We conduct extensive token pruning experiments for the widely used vision transformer backbones, DeiT \cite{touvron2021training} and LV-ViT \cite{jiang2021all} on ImageNet. For instance, our method improves 1.5x throughput with only 0.2\% decrease in accuracy while reducing the 35\% GFLOPs of the DeiT-small model. For other model and pruning rates, our method also achieves a better accuracy and complexity balance compared with previous approaches.%


\section{Related Work}
\paragraph{Vision Transformer.} Transformer \cite{transformer}, developed in NLP, has been successfully applied to various vision tasks and tends to replace CNN \cite{he2016deep,tan2019efficientnet} gradually. ViT \cite{dosovitskiy2020image} illustrates that transformer lacks inductive bias and requires large-scale dataset pre-training to achieve an approximate performance with the state-of-the-art CNN. DeiT \cite{touvron2021training} eliminates the above problem with well-tuned training parameters and the introduction of distillation token. LV-ViT \cite{jiang2021all} explores a variety of techniques for training vision transformer, significantly improving the performance of ViT. PVT \cite{wang2021pyramid} constructs a hierarchical ViT similar to CNN and proposes spatial-reduction attention to reduce the computation complexity. Swin Transformer \cite{liu2021swin} proposes a window based self-attention that makes the complexity linear with respect to token number and becomes a generic visual backbone. DGT~\cite{liu2022dynamic} propose the dynamic group attention to accelerate inference.

\paragraph{Static ViT Pruning.} Analogous to weights pruning in CNN \cite{han2015deep,li2016pruning,molchanov2016pruning,Liu2017learning,luo2017thinet,frankle2018lottery}, there are many works \cite{zhu2021vision,yang2021nvit,chen2021chasing,yu2022unified} for ViT parameters compression. VTP \cite{zhu2021vision} prunes parameters in $\MHSA$ and $\FFN$ indiscriminately through L1 regularized sparse training. NViT \cite{yang2021nvit} establishes the global weights importance via performing Taylor expansion to the loss, then conducts structured pruning and parameter reassignment based on dimensional trends. SViTE \cite{chen2021chasing} fully explores the sparsity of ViT, compressing transformer with structured pruning, unstructured sparsity and token pruning. In this paper, we focus on token compression, which is orthogonal to static ViT pruning, and we can further improve the compression rate, benefit from weights pruning.

\paragraph{Dynamic ViT Pruning.} Thanks to the transformer's parallel computing mechanism, pruning image tokens can bring real acceleration without the need of additional support. Both DynamicViT \cite{rao2021dynamicvit} and IA-RED$^2$ \cite{pan2021ia} dynamically keep informative tokens by inserting prediction modules. PS-ViT \cite{tang2021patch} belongs to static token pruning, which statistically obtains the importance distribution of tokens across the dataset and prunes them top-down. EViT \cite{liang2022evit} and Evo-ViT \cite{xu2022evovit} use the class attention score to distinguish how informative a token is. A-ViT \cite{yin2021adavit} adaptively calculates the halting score for each token. Our method strives to implement instance-wise token pruning without additional computational costs.

\section{Method}

\subsection{Preliminary}
\label{bk}
Vision Transformer \cite{dosovitskiy2020image} provides a new paradigm for image recognition. ViT first splits the image into $N \times N$ non-overlapping patches and embeds them into a $D$ dimensional feature space, and then adds a class token before the image tokens for final classification. Considering the position relationship, all tokens are added with a learnable position encoding and then fed into a stacked transformer block. We summarize the operations inside the block by using the following two equations:
\begin{gather}
    x_{\text{MHSA}} = x + \MHSA(\LN(x)), \\
    x_{\text{FFN}} = x_{\text{MHSA}} + \FFN(\LN(x_{\text{MHSA}})),
\end{gather}
where $\MHSA$ stands for Multi-Head Self-Attention \cite{transformer}, $\FFN$ is feed-forward neural network, and $\LN$ stands for layer normalization.

$\MHSA$ represents features in multiple subspaces using the same parameters. The input is first projected into three matrices Query $\mathbf{Q} \in \mathbb{R}^{(N+1)\times D}$, Key $\mathbf{K} \in \mathbb{R}^{(N+1)\times D}$, and Value $\mathbf{V} \in \mathbb{R}^{(N+1)\times D}$, respectively, and then sliced into $H$ attention heads to perform the parallel operations:
\begin{align}
    \CTX(\mathbf{Q},\mathbf{K},\mathbf{V}) = \softmax
    \left( \frac{\mathbf{Q}\mathbf{K}^T}{\sqrt{D_{h}}}\right)\mathbf{V},
\label{eq4}
\end{align}
where $D_{h}=\frac{D}{H}$ is the feature dimension of the single head output. In particular, the attention scores of class token $x_{\text{cls}}$ and other tokens can be written as:
\begin{equation}
    \ascore(x_{\text{cls,:}}) = \softmax\left(\frac{\mathbf{Q}_{\text{cls}}\mathbf{K}^T}{\sqrt{D_{h}}}\right) \in \mathbb{R}^{H \times N},
\end{equation}
which is used to reflect which tokens are contributing to the classification.
Next, $\MHSA$ concatenates the $\CTX$ of all attention heads together and project them through a matrix.

\subsection{Class Attention Score Weighted by Attention Head Importance}
\label{ts}

Popular metrics for evaluating the importance of a token include its similarity score to the class token~\cite{liang2022evit,xu2022evovit} and the attention it receives from other tokens~\cite{goyal2020power,kim2021learned}. However, we observed that both approaches ignore the diversity of attention heads, and the scores of different heads are treated equally, in other words, the final score is their average value.

\begin{figure}[t]
\centering
\includegraphics[width=\columnwidth]{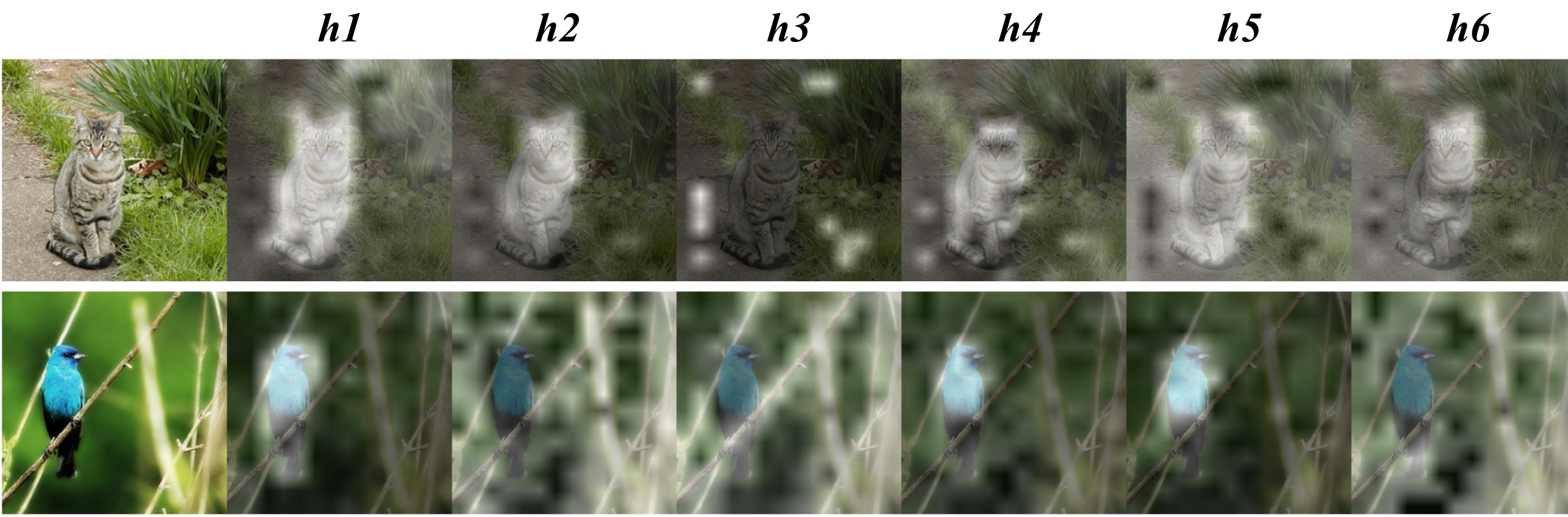}
\caption{Visualization of tokens' attention paid to each head. The figure on the top is the token-level attention head importance visualization from AS-DeiT-S layer 7, and the figure below is from AS-DeiT-S layer 9.}
\label{fig:head}
\end{figure}

\begin{figure*}[t]
\centering
\includegraphics[width=0.8\textwidth]{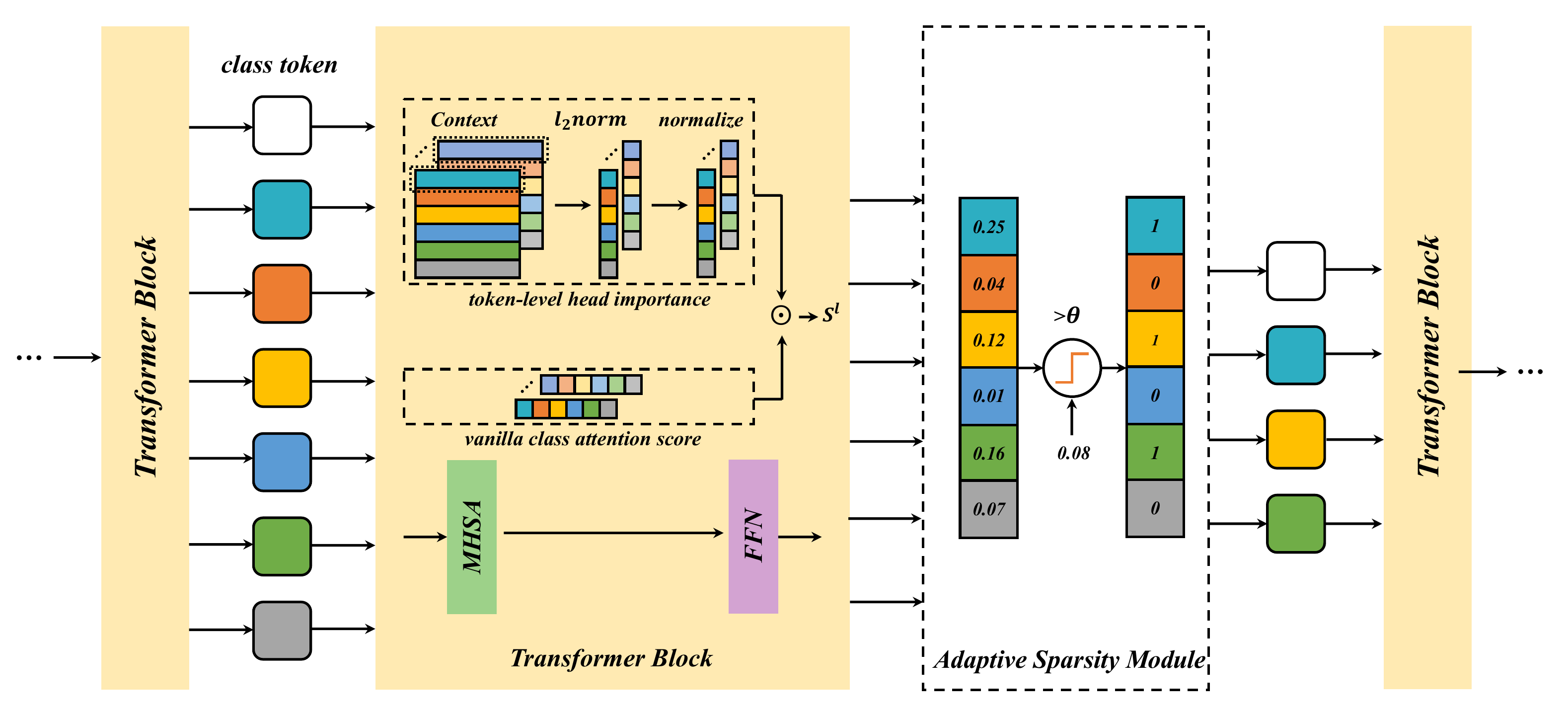}
\caption{The framework of our Adaptive Sparse ViT. $\theta$ denotes the learnable threshold and $\odot$ indicates Hadamard product. The token evaluation metric accounts for the attention of class token as well as the importance of different attention heads. Adaptive Sparse Module fulfills pruning by comparing token scores and thresholds.}
\label{fig:frame}
\end{figure*}

Considering that different tokens may receive diverse attention in multiple attention heads, we propose a metric to estimate the token-level head importance, and it simply relies on the intermediate results of $\MHSA$. Taking the $i$-th input token $x_i$ as an example, firstly, we define the $\CTX$ of the $h$-th head of the $l$-th layer as $\CTX^{(h,l)}(x_i) \in \mathbb{R}^{D_{h}}$, then we calculate the $l_{2}$-norm along the last dimension as the importance $\mathcal{H}^{(h,l)}(x_i)$ of the $h$-th head like Equation (\ref{eq6}):
\begin{align}
    \label{eq6}
    \mathcal{H}^{(h,l)}(x_i)=\sqrt{\sum_{j=1}^{D_{h}}{\CTX_{j}^{(h,l)}(x_i)}^2}.
\end{align}
Our motivation for estimating head importance derives from CNN pruning, of which $l_{2}$-norm \cite{he2018soft} is a commonly used metric for filter significance rating. In addition, our metrics are token-level, different from the previous head pruning work \cite{michel2019sixteen}. Next, we take the proportion of the importance of the $h$-th attention head to the sum of all head as its weighting factor like Equation (\ref{eq:r}):
\begin{gather}
    \label{eq:r}
    \mathcal{R}^{(h,l)}(x_i)=\frac{\mathcal{H}^{(h,l)}(x_i)}{\sum_{h=1}^{H}\mathcal{H}^{(h,l)}(x_i)}, \\
    \label{eq:s}
    \mathcal{S}^{l}(x_i)=\sum_{h=1}^{H}\mathcal{R}^{(h,l)}(x_i) \cdot \ascore^{(h,l)}(x_{\text{cls,i}}), \quad i=1,2,...,N.
\end{gather}

We visualize the token-level head importance in the Figure \ref{fig:head} and brighter areas indicate that the current attention head is more important for this token. Taking the second image as an example, the foreground tokens focus more on head 1, 4 and 5, while the background tokens clearly favor head 2 and 3. It is clear that the diversity of individual heads should be taken into account when scoring tokens.

The vanilla class attention score $\ascore^{(h,l)}(x_{\text{cls,i}})$ of the $i$-th token is chosen as the basic evaluation metric, and we can estimate token score $\mathcal{S}^{l}(x_i)$ more accurately through head importance weighting as Equation (\ref{eq:s}). The flow of head importance weighted class attention score is shown in Figure \ref{fig:frame}. Compared to existing adaptive token pruning works, our approach takes into account the attention head diversity and has no reliance on any extra scoring mechanisms.

\subsection{Adaptive Token Pruning Based on Learnable Thresholds} 
In order to achieve sample-adaptive token pruning while minimizing the operation cost. We introduce three stage-wise learnable thresholds to control token sparse behavior, following LTP \cite{kim2021learned}. Figure \ref{fig:frame} illustrates the overall framework of our proposed method. We usually divide the successive transformer blocks into four stages and insert a learnable threshold $\theta$ before the second, third and fourth stages, respectively. Adaptive Sparse Module is a comparator based on thresholds and token scores. We keep tokens with scores greater than the threshold, as in Equation (\ref{eq5}):
\begin{equation}
    M^{n}(x_i) = \begin{cases}
      &1, \quad \text{ if } \mathcal{S}^{l}(x_i) >\theta^{n} \\
      &0, \quad \text{otherwise} 
    \end{cases}, \quad n=1,2,3,
\label{eq5}
\end{equation}
where $M^{n}(x_i)$ is a binary mask to indicate whether the $i$-th token is pruned or not, and $\mathcal{S}^{l}(x_i)$ denotes token score of the layer before the $n$-th stage. The token scores corresponding to different input images usually present different distributions, so using a threshold to truncate the distribution can yield a specific pruning strategy for each instance. 

Considering images in the same mini-batch have different pruning configurations, for the sake of parallel training, we cannot simply discard uninformative tokens during the training process. $M^{n}(x_i)$ can be used to explicitly cut the connection between uninformative tokens and useful tokens. There are two masking methods, attention mask~\cite{rao2021dynamicvit} and activation mask \cite{kim2021learned}, the former acting on the attention matrix and the latter on the output of the $\FFN$.


However, it is not easy to optimize the learnable thresholds. The pruning mask comes from comparison, blocking the gradient back propagation, making the threshold untrained. We transform the hard mask into a soft differentiable mask using the sigmoid function:
\begin{equation}
    \tilde{M}^{n}(x_i) = \sigmoid(T\cdot (\mathcal{S}^{l}(x_i)-\theta^{n})),\quad n=1,2,3.
\end{equation}
To approximate the hard mask, we employ a temperature parameter $T$, where the $\sigmoid$ function behaves close to the step function at a sufficiently high temperature. The soft mask is differentiable, and by using the gradient straight through estimator (\textbf{STE}), we are able to optimize the learnable threshold as normal. A detailed implementation is given in the Algorithm \ref{alg:code}.

\subsection{Budget-Aware Training}

We achieve token pruning by constraining the target computation across the dataset. Compared to DynamicViT \cite{rao2021dynamicvit}, which manually sets the token sparsity ratio at each stage to indirectly control the complexity, our method automatically achieves a good trade-off between accuracy and complexity given only a budget. Specifically, we propose a budget-aware loss ($\mathcal{L}_{\text{FLOPs}}$). Given a mini-batch inputs $x$ with size of $B$, we can obtain their average FLOPs and then make the actual computational cost close to the target budget using MAE loss as Equation (\ref{eq11}):
\begin{equation}
    \mathcal{L}_{\text{FLOPs}} = \left \| \flops(x, \Theta)/B-t  \right \|_{1},
\label{eq11}
\end{equation}
where $\flops$ is a function to calculate the actual operations for different inputs under the effect of all thresholds $\Theta$, and $t$ is the expected complexity. The intention of designing the budget-aware loss is consistent with our sample-adaptive approach. The learnable thresholds are optimized to the appropriate range in a data-driven manner under the budget constraint, without imposing other artificial restrictions.

Our training objective include other two parts. The first part is the regular cross-entropy loss ($\mathcal{L}_{\text{CE}}$) as following equation:
\begin{equation}
    \mathcal{L}_{\text{CE}} = \CE(y,\hat{y})
\end{equation}
where $y$ denotes the ground truth labels and $\hat{y}$ the $\softmax$ output. To further improve the performance, we consider transferring the knowledge of the full model to the compressed network during training. Let the distribution $\hat{y_{t}}$ denote the prediction of the teacher network, and we use KL divergence to minimize the gap between the output of the student network and the teacher network. For vision transformers like LV-ViT \cite{jiang2021all} with an additional linear layer to integrate all image tokens, it also needs to be aligned with the teacher network. Therefore, distillation loss ($\mathcal{L}_{\text{distill}}$) can be written as:
\begin{equation}
    \mathcal{L}_{\text{distill}} = \kld(\hat{y}, \hat{y_{t}}) \quad \text{or} \quad \mathcal{L}_{\text{distill}} = \kld(\hat{y}, \hat{y_{t}}) + \kld(\hat{z}, \hat{z_{t}}).
\end{equation}
The overall training objective is a combination of the above three components:
\begin{equation}
    \mathcal{L} = \mathcal{L}_{\text{CE}} + \lambda_{1}\mathcal{L}_{\text{FLOPs}} + \lambda_{2}\mathcal{L}_{\text{distill}},
\end{equation}
where $\lambda$ is used to control the loss balance, and we set $\lambda_{1}=2$, $\lambda_{2}=0.5$ in our experiments.

\begin{table*}[t]
  \centering
    \begin{tabu}to\textwidth{l*{5}{X[c]}}\toprule
    Model & Params (M) & GFLOPs & Top-1 Acc (\%) & Throughput (img/s) & Latency (ms) \\\midrule
    DeiT-S~\cite{touvron2021training} & 22.1  & 4.6  & 79.8  & 770  & 6.16  \\ \midrule
    DyViT/$\rho$=0.7~\cite{rao2021dynamicvit} & 22.8  & 2.9  & 79.3 (-0.5)  & 1155 (+50\%)  & 7.95 (+29\%) \\
    PS-ViT~\cite{tang2021patch} & 22.1  & 2.6  & 79.4 (-0.4)  & -  & - \\
    IA-RED$^2$~\cite{pan2021ia} & -  & 3.2  & 79.1 (-0.7)  & -  & - \\
    Evo-ViT~\cite{xu2022evovit} & 22.1  & 3.0  & 79.4 (-0.4)  & 1143 (+48\%)  & 8.66 (+41\%) \\
    EViT-DeiT-S/$\rho$=0.7~\cite{liang2022evit} & 22.1  & 3.0  & 79.5 (-0.3) & 1149 (+49\%)  & 7.3 (+19\%) \\
    A-ViT~\cite{yin2021adavit} & 22.1  & 3.6  & 78.6 (-1.2)  & -  & - \\
    \textbf{AS-DeiT-S/$f$=0.65} (Ours) & 22.1  & 3.0  & 79.6 \cb{(-0.2)}  & 1192 \cb{(+55\%)}  & 6.56 \cb{(+6\%)} \\
    \midrule
    DyViT/$\rho$=0.5~\cite{rao2021dynamicvit} & 22.8 & 2.2 & 77.5 (-2.3) & -  & - \\
    EViT-DeiT-S/$\rho$=0.5~\cite{liang2022evit} & 22.1  & 2.3  & 78.5 (-1.3)  & 1494 (+94\%)  & 7.19 (+17\%) \\
    \textbf{AS-DeiT-S/$f$=0.5} (Ours) & 22.1  & 2.3  & 78.7  \cb{(-1.1)}  & 1520 \cb{(+97\%)}  & 6.38 \cb{(+4\%)} \\\midrule \midrule
    LV-ViT-S\cite{jiang2021all} & 26.2  & 6.6  & 83.3  & 571  & 9.24 \\ \midrule
    DyViT-LV-S/$\rho$=0.7~\cite{rao2021dynamicvit} & 26.9  & 4.6  & 83.0 (-0.3)  & 807 (+41\%)  & 11.06 (+20\%) \\
    EViT-LV-S/$\rho$=0.7~\cite{liang2022evit} & 26.2  & 4.7  & 83.0 (-0.3)  & -  & - \\
    \textbf{AS-LV-S/$f$=0.7} (Ours) & 26.2  & 4.6  & 83.1 \cb{(-0.2)}  & 884 \cb{(+55\%)}  & 9.20 \cb{(-0\%)} \\\midrule
    DyViT-LV-S/$\rho$=0.5~\cite{rao2021dynamicvit} & 26.9  & 3.7  & 82.0 (-1.3)  & 1011 (+77\%)  & 11.04 (+19\%) \\
    EViT-LV-S/$\rho$=0.5~\cite{liang2022evit} & 26.2  & 3.9  & 82.5 (-0.8)  & -  & - \\
    \textbf{AS-LV-S/$f$=0.6} (Ours) & 26.2  & 3.9  & 82.6 \cb{(-0.7)}  & 1023 \cb{(+80\%)}  & 9.31 \cb{(+1\%)} \\\midrule \midrule
    LV-ViT-M\cite{jiang2021all} & 55.8  & 12.7  & 84.1  & 317  & 11.63 \\ \midrule
    DyViT-LV-M/$\rho$=0.8~\cite{rao2021dynamicvit} & 57.1  & 9.6  & 83.9 (-0.2)  & -  & - \\
    \textbf{AS-LV-M/$f$=0.76} (Ours) & 55.8  & 9.6  & 84.0 \cb{(-0.1)}  & 657 \cb{(+107\%)}  & 11.43 \cb{(-2\%)} \\\midrule
    DyViT-LV-M/$\rho$=0.7~\cite{rao2021dynamicvit} & 57.1  & 8.5  & 83.8 (-0.3)  & 476 (+50\%)  & 13.29 (+14\%) \\
    \textbf{AS-LV-M/$f$=0.67} (Ours) & 55.8  & 8.5  & 83.9 \cb{(-0.2)}  & 801 \cb{(+153\%)}  & 11.53 \cb{(-1\%)}
    \\\bottomrule
    \end{tabu}%
    \caption{Comparisons with the previous token pruning methods. Use $\rho$ to denote the token keeping rate and $f$ to indicate the proportion of budget. GFLOPs represents the average computational costs across the dataset. The throughput metric is measured on a single NVIDIA 2080Ti GPU using a fixed batch size 64 and hardware latency is the average elapsed time of 100 inferences with a single image on the same machine.}
  \label{tab:compare}%
\end{table*}%

\paragraph{Inference.} Learnable thresholds are fixed after the training. Given an input image to do inference, we only need the intermediate result of $\MHSA$ to get the token score effortlessly. The pruning process is also simple enough that our method only needs one comparison to know which tokens are to be kept, saving the topK computation cost compared to the previous work~\cite{rao2021dynamicvit,xu2022evovit,liang2022evit}.

\section{Experiments}
\subsection{Implementation Details}
Our experiments are conducted on the ImageNet-1K \cite{deng2009imagenet} classification dataset, with token pruning performed on the popular DeiT \cite{touvron2021training} and LV-ViT \cite{jiang2021all} models. We finetune the pre-trained model by 30 epoch to obtain the compressed network, and most of the training settings stay the same as the originals. More experiment details can be found in the Appendix \ref{sec:detail}.

\begin{table}[t]
\centering
\begin{tabular}{c|cc|c|c}
\toprule
Model     & ASM & HS & Acc (\%) & \makecell[c]{Throughput \\ (img/s)} \\ \midrule
          &     &      & 79.36   & 1099 \\
          &     & \checkmark    & 79.36  & 1086  \\
          & \checkmark   &      & 79.56 \cb{(+0.2)}  & 1198 \\
AS-DeiT-S & \checkmark   & \checkmark    & 79.63 \cb{(+0.27)}   & 1192 \\ 
\bottomrule
\end{tabular}
\caption{Effectiveness of each module.}
\label{tab:ablation_a}
\end{table}

\begin{table}[t]
\centering
\begin{tabular}{lcc}
    \toprule
        Method & Acc (\%)  & GFLOPs \\ \midrule
        + attention\_mask & 79.63  & 3.0 \\
        + activation\_mask & 78.04 \cb{(-1.59)}  & 3.0 \\
        w/o $\mathcal{L}_{\text{distill}}$ & 79.46 \cb{(-0.17)} & 3.0 \\
    \bottomrule
\end{tabular}
\caption{Effect of different masking strategies and distillation.}
\label{tab:ablation_b}
\end{table}

\begin{table}[t]
  \centering
  \resizebox{1.0\columnwidth}{!}{
    \begin{tabular}{lc|cccc}
    \toprule
      &Method     & $\rho$=0.9 & $\rho$=0.8 & $\rho$=0.7 & $\rho$=0.5 \\ \midrule
    pretrained  &vanilla  &79.77	&79.24	&78.51	&73.72 \\ \midrule
    pretrained  &HS    &79.79	&79.29	&78.51	&73.78   \\ \midrule
    finetuned  &AS-ViT      & 79.8   & 79.7     & 79.6   & 78.7 \\
    \bottomrule
    \end{tabular}
}
\caption{Comparison with the raw attention of the pre-trained model.}
\label{tab:raw}
\end{table}

\subsection{Main Results}
\paragraph{Performance Comparisons With Existing Pruning Methods.} We test the Top-1 accuracy, throughput and hardware latency of three models, DeiT-S, LVViT-S and LVViT-M, under different pruning budgets and compare them with previous token pruning work. The experimental results on accuracy are illustrated in Table~\ref{tab:compare}. Compared to previous work, our proposed adaptive sparse ViT achieves state-of-the-art performance with similar complexities. For all models, the Top-1 accuracy degradation of our pruned models is controlled within 0.2\% when the computation decreases by 30\%$\sim$35\%. When the compression rate of DeiT-small is further increased to 50\%, the advantage of sample-adaptive token sparsity becomes more obvious over the fixed pruning rate approaches~\cite{rao2021dynamicvit,liang2022evit}, as they may be forced to discard some important tokens. Moreover, our method is far better than the sample-adaptive \cite{pan2021ia,yin2021adavit}, owing to the more accurate token scoring mechanism. The reinforcement learning strategy employed by IA-RED$^2$~\cite{pan2021ia} is difficult to converge.

\paragraph{Efficiency Comparisons With Existing Pruning Methods.} The experimental results on efficiency are in Table~\ref{tab:compare}. Note that some previous methods did not release code and trained weights, so we cannot compare and report them. Compared to previous work, AS-ViT achieves the best results in terms of accuracy, throughput and hardware latency. In contrast to methods like DynamicViT \cite{rao2021dynamicvit} that use extra modules to calculate token scores, AS-ViT relies only on the intermediate results of $\MHSA$ to evaluate tokens, thus saving expensive computations and achieving higher throughput metrics. The threshold-based comparator employed by Adaptive Sparsity Module saves the computational cost of ranking all tokens, which makes our latency metrics significantly better than previous work.

\paragraph{Comparisons With Other Models.} We compare the complexity and accuracy of our adaptive sparse LV-ViT (abbreviated as AS-LV-ViT) with other sota models on ImageNet in Figure \ref{fig:sota}. Detailed data can be found in Table \ref{tab:sota}. Our AS-LV-ViT shows quite competitive performance under different complexities, with far higher throughput than other CNN~\cite{tan2019efficientnet,liu2022convnet} and ViT~\cite{chu2021twins,wang2021pyramid,xu2021co,wu2021cvt,touvron2021cait,jiang2021all,Yuan_2021_ICCV} while still providing advanced accuracy. In addition, by simply adjusting the budget, our approach achieves a better accuracy-complexity trade-off compared automatically, avoiding the tedium of manual design.

\subsection{Ablation Analysis}
\paragraph{Effectiveness of Each Sub-Module.} 
In Table \ref{tab:ablation_a}, we study the effect of each module in detail. \textbf{ASM} represents the Adaptive Sparsity Module, and \textbf{HS} is attention head importance weighted class attention score. In the ablation experiments, we use fixed ratio topK module instead of ASM and vanilla class attention score to replace HS. It is obvious that the Adaptive Sparsity module significantly improves the model performance compared with the fixed ratio module by 0.2\%, which fully illustrates the necessity and effectiveness of sample adaptive token pruning. With ASM, the attention head weighted class attention score improves the precision by 0.07\% compared to the original metric without a noticeable reduction in throughput. In addition, we can observe that HS needs to be used with ASM to have better results.

\paragraph{Effectiveness of Training Techniques.} We apply masking strategy to achieve parallel training and knowledge distillation to stabilize the training process. The respective experimental results are listed in Table \ref{tab:ablation_b}. We conduct experiments with DeiT-Small. Obviously, attention\_mask is far better than activation\_mask, which can shield the uninformative tokens from interacting with other tokens. And by transferring the knowledge of full model to the compressed model, we can further improve the accuracy.

\paragraph{Comparison With Raw Attention Baseline of Pre-Trained Model.} In this part, we consider the attention of the pre-trained model as a token evaluation metric and discard unimportant tokens in a fixed proportion. In addition, we apply head importance weighted class attention score (\textbf{HS}) directly on the raw attention baseline to fully demonstrate the effectiveness of our method. The data in Table \ref{tab:raw} present the accuracy of each method at different keeping rates. The attention map of the pre-trained model is no longer reliable and the accuracy decreases significantly as the compression ratio gradually increases. While our method can still achieve the similar performance of the original model by fine-tuning. Furthermore, our head importance weighted token score can produce positive results without training.

\paragraph{Batch Inference.} Our approach achieves both sample-adaptive and batch-adaptive token pruning. The fact that AS-ViT is well suited for single-image computation scenarios does not mean that it cannot be used for parallel inference. Since we use the average complexity of mini-batch to calculate the budget-aware loss, we can also do parallel inference from the mean value of the number of kept tokens within a batch. And the experimental results in the Table \ref{tab:bs} show that this does not cause significant accuracy degradation.

\begin{table}[t]
\centering
\begin{tabular}{ccccc}
    \toprule
    Batch Size     & 1 & 32 & 64 & 128 \\ \midrule
    Acc (\%)    & 79.63    & 79.60     & 79.58    & 79.61   \\ \midrule
    GFLOPs      & 3.0    & 3.0     & 3.0   & 3.0   \\
    \bottomrule
\end{tabular}
\caption{Accuracy on ImageNet with different batch\_size.}
\label{tab:bs}
\end{table}

\begin{figure}[t]
\centering
\includegraphics[width=0.96\columnwidth]{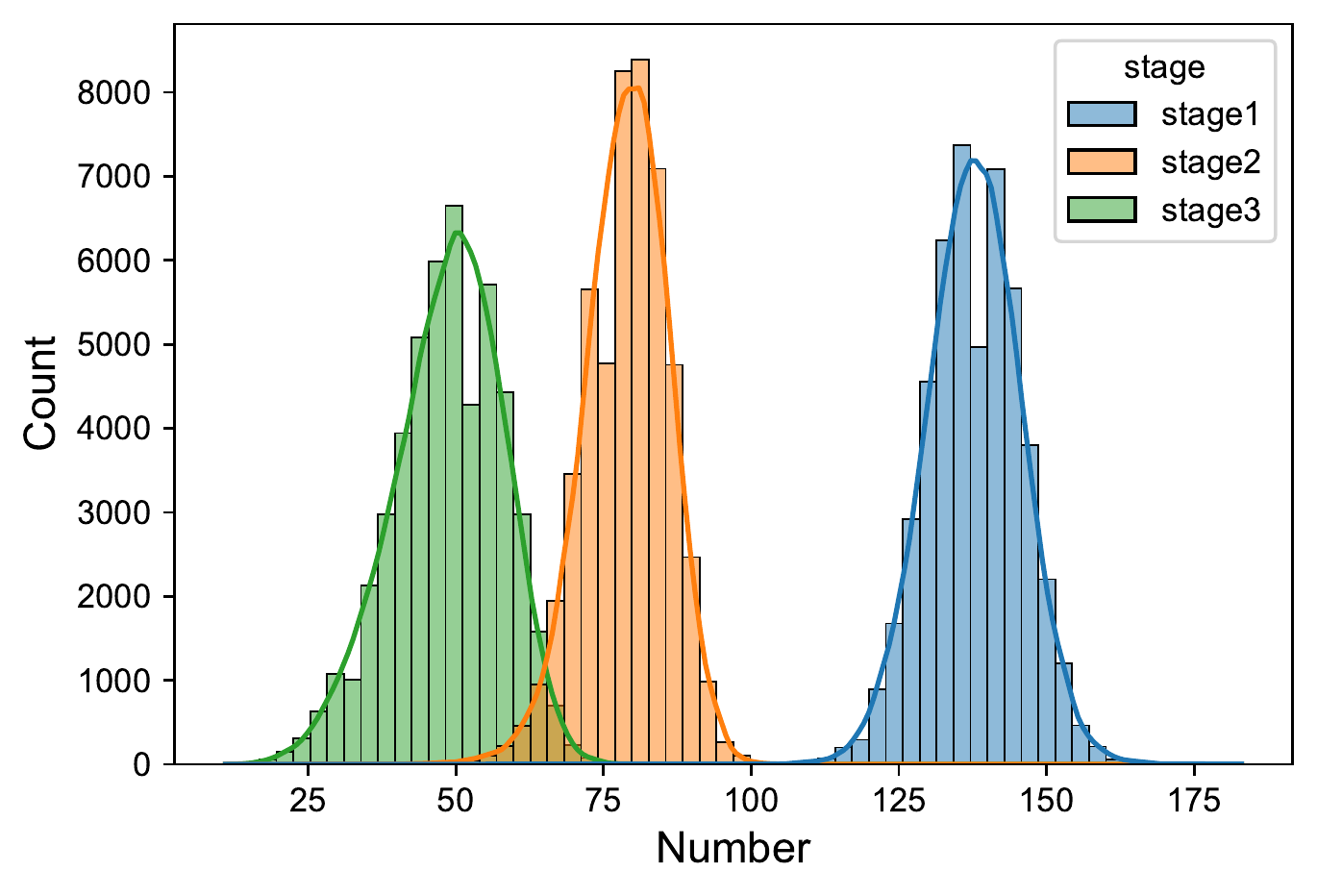}
\caption{Distribution of the token number at different stages.}
\label{fig:distribution}
\end{figure}

\begin{table*}[t]
  \centering
    \begin{tabu}to\textwidth{c*{4}{X[c]}}\toprule
    Location & GFLOPs & Top-1 Acc (\%) & Throughput (img/s) & Latency (ms) \\\midrule
    $\mathbf{[4, 7, 10]}$ & 3.0  & \textbf{79.63}  & 1192  & 6.56  \\
    $[3, ~6, ~9]$  & 3.0  & 79.46  & 1184  & 6.54 \\
    $[3, ~5, 7, ~9]$  & 3.0  & 79.46  & 1191  & 6.60 \\
    $[4, 6, 8, 10]$  & 3.0  & 79.5  & 1195  & 6.65 \\
    $[3,4,...,10,11]$  & 3.0  & 78.8  & 1146  & 7.6 \\\bottomrule
    \end{tabu}%
    \caption{Impacts of module insertion configurations on accuracy and speed. We use the DeiT-small model with 65\% FLOPs as the baseline.}
  \label{tab:loc}%
\end{table*}%

\begin{table}[t]
\centering
\resizebox{\columnwidth}{!}{
\begin{tabular}{lccc}
    \toprule
    Method     & Resolution	  & Acc (\%) & GFLOPs  \\ \midrule
    DeiT-B~\cite{touvron2021training}    & 224     & 81.8   & 17.5  \\ \midrule
    DyViT-B~\cite{rao2021dynamicvit}  & 224      & 81.3 (-0.5)  & 11.2  \\
    EViT-DeiT-B~\cite{liang2022evit}  & 224     & 81.3 (-0.5)   & 11.5  \\
    IA-RED2~\cite{pan2021ia}   & 224      & 80.3 (-1.5)   & 11.8  \\
    AS-DeiT-B   & 224      & 81.4 \cb{(-0.4)}  & 11.2  \\
    \midrule \midrule
    DeiT-B~\cite{touvron2021training}   & 384       & 82.9  & 49.4  \\ \midrule
    IA-RED2~\cite{pan2021ia}   & 384       & 81.9 (-1.0)  & 34.7  \\
    AS-DeiT-B   & 384      & 82.7 \cb{(-0.2)}  & 34.6  \\
    \bottomrule
\end{tabular}
}
\caption{Experimental results of token pruning in large model and at different input resolutions.}
\label{tab:big}
\end{table}

\paragraph{Pruning Location.} Referring to previous works~\cite{rao2021dynamicvit,liang2022evit}, we adopts a progressive token sparse strategy. The position and number of our Adaptive Sparsity Module are need to consider. We experiment with several insertion configurations to demonstrate that the current strategy is accuracy-latency optimal. The results are given in Table \ref{tab:loc}. The current method has higher accuracy compared to the [3,6,9] pruning strategy. This may be attributed to the early layer's class attention score not being stable enough, causing some information tokens to be discarded incorrectly. When increasing to 4 modules, there is no significant change in throughput and latency. We further insert thresholds from the 3rd to 11th layers, and the training becomes unstable and brings severe accuracy degradation, which we speculate is attributed to the difficulty of optimizing multiple thresholds in the limited fine-tuning. In addition, frequent reorganization of tokens causes non-negligible time consumption.

\paragraph{Performance on Large Model and Different Resolutions.} To fully demonstrate the effectiveness of our method, we perform token sparse on the large baseline model DeiT-Base at different input resolutions. As illustrated in the Table \ref{tab:big},  AS-ViT performs better than previous work under the same resolution and complexity. When using larger resolutions, our method is significantly better than IA-RED$^2$ \cite{pan2021ia}. Furthermore, at a resolution of 384x384, the accuracy degradation is smaller compared to 224x224, suggesting that there is greater redundancy at higher resolutions, which can improve efficiency through token pruning.

\begin{figure}[t]
\centering
\includegraphics[width=0.92\columnwidth]{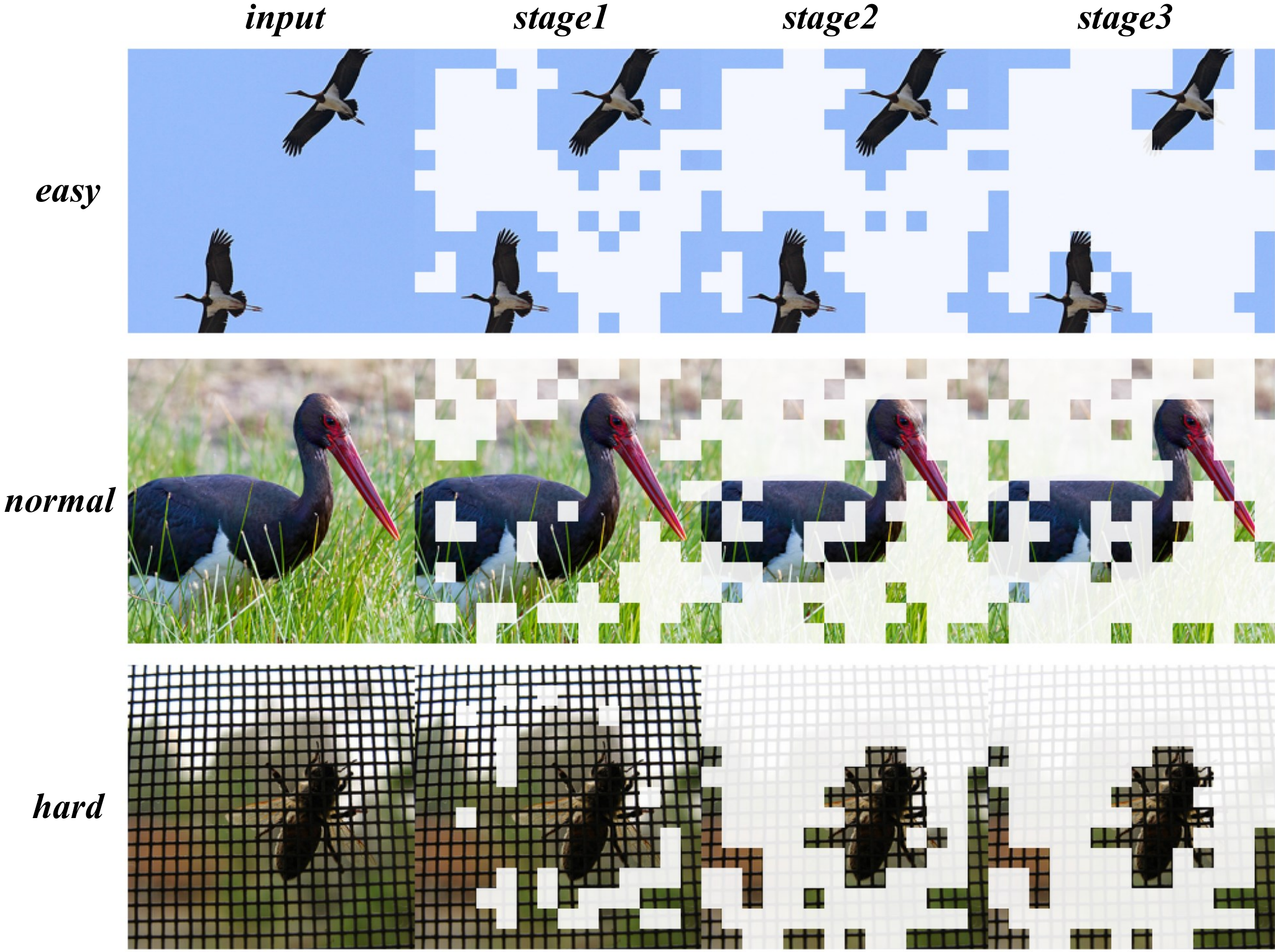}
\caption{Visualization results of token pruning for samples with different recognition difficulties.}
\label{fig:viz}
\end{figure}

\paragraph{Visualization.} To analyze the pruning behavior of our adaptive token sparse method on different images, we count the amount of kept tokens of the AS-DeiT-S model in each stage for ImageNet validation dataset images and plot their distribution in Figure \ref{fig:distribution}. The number of each stage basically shows a Gaussian distribution, peaking at a certain value. This suggests that a fixed proportion of pruning is also relatively reasonable, since the easy and difficult samples are only a minority. However, our adaptive token sparse method can give the corresponding pruning configurations for samples with different recognition difficulties, which is reflected in the distribution extending to both sides. Further, we select three representative images for visualization of token pruning results in Figure \ref{fig:viz}. For easy samples, AS-ViT discards plenty of useless tokens in the early stage to save computational cost, While for complicated and hard to recognize instances, the model tends to discard more tokens in later stages with higher confidence. More image token pruning results can be found in Figure \ref{fig:imgviz}.

\section{Conclusion}
In this work, we propose a sample-adaptive token pruning method, which effortlessly evaluates token importance via fully exploiting the MHSA mechanism, and then introduces learnable thresholds to accomplish pruning. Our proposed budget-aware loss can effectively constrain the thresholds so that the pruned model reaches the complexity budget. Compared to previous work, our approach does not require additional sub-networks to compute token scores and also performs specific pruning strategies for different samples just by comparing with thresholds. Experimental results on various models show that our AS-ViT greatly improves the throughput and achieves lower latency without significantly affecting the accuracy. In the future, we will transfer token pruning to downstream tasks and combining it with static parameters pruning for further acceleration.

\bibliographystyle{named}
\bibliography{ijcai23}

\begin{thebibliography}{}

\bibitem[\protect\citeauthoryear{Carion \bgroup \em et al.\egroup
  }{2020}]{carion2020end}
Nicolas Carion, Francisco Massa, Gabriel Synnaeve, Nicolas Usunier, Alexander
  Kirillov, and Sergey Zagoruyko.
\newblock End-to-end object detection with transformers.
\newblock In {\em ECCV}, pages 213--229, 2020.

\bibitem[\protect\citeauthoryear{Chen \bgroup \em et al.\egroup
  }{2021}]{chen2021chasing}
Tianlong Chen, Yu~Cheng, Zhe Gan, Lu~Yuan, Lei Zhang, and Zhangyang Wang.
\newblock Chasing sparsity in vision transformers: An end-to-end exploration.
\newblock {\em Advances in Neural Information Processing Systems}, 34, 2021.

\bibitem[\protect\citeauthoryear{Chu \bgroup \em et al.\egroup
  }{2021}]{chu2021twins}
Xiangxiang Chu, Zhi Tian, Yuqing Wang, Bo~Zhang, Haibing Ren, Xiaolin Wei,
  Huaxia Xia, and Chunhua Shen.
\newblock Twins: Revisiting the design of spatial attention in vision
  transformers.
\newblock {\em Advances in Neural Information Processing Systems}, 34, 2021.

\bibitem[\protect\citeauthoryear{Deng \bgroup \em et al.\egroup
  }{2009}]{deng2009imagenet}
Jia Deng, Wei Dong, Richard Socher, Li-Jia Li, Kai Li, and Li~Fei-Fei.
\newblock Imagenet: A large-scale hierarchical image database.
\newblock In {\em CVPR}, pages 248--255, 2009.

\bibitem[\protect\citeauthoryear{Dosovitskiy \bgroup \em et al.\egroup
  }{2020}]{dosovitskiy2020image}
Alexey Dosovitskiy, Lucas Beyer, Alexander Kolesnikov, Dirk Weissenborn,
  Xiaohua Zhai, Thomas Unterthiner, Mostafa Dehghani, Matthias Minderer, Georg
  Heigold, Sylvain Gelly, et~al.
\newblock An image is worth 16x16 words: Transformers for image recognition at
  scale.
\newblock In {\em International Conference on Learning Representations (ICLR)},
  2020.

\bibitem[\protect\citeauthoryear{Frankle and Carbin}{2018}]{frankle2018lottery}
Jonathan Frankle and Michael Carbin.
\newblock The lottery ticket hypothesis: Finding sparse, trainable neural
  networks.
\newblock In {\em International Conference on Learning Representations}, 2018.

\bibitem[\protect\citeauthoryear{Goyal \bgroup \em et al.\egroup
  }{2020}]{goyal2020power}
Saurabh Goyal, Anamitra~Roy Choudhury, Saurabh Raje, Venkatesan Chakaravarthy,
  Yogish Sabharwal, and Ashish Verma.
\newblock Power-bert: Accelerating bert inference via progressive word-vector
  elimination.
\newblock In {\em International Conference on Machine Learning}, pages
  3690--3699. PMLR, 2020.

\bibitem[\protect\citeauthoryear{Han \bgroup \em et al.\egroup
  }{2015}]{han2015deep}
Song Han, Huizi Mao, and William~J Dally.
\newblock Deep compression: Compressing deep neural networks with pruning,
  trained quantization and huffman coding.
\newblock {\em arXiv preprint arXiv:1510.00149}, 2015.

\bibitem[\protect\citeauthoryear{He \bgroup \em et al.\egroup
  }{2016}]{he2016deep}
Kaiming He, Xiangyu Zhang, Shaoqing Ren, and Jian Sun.
\newblock Deep residual learning for image recognition.
\newblock In {\em CVPR}, pages 770--778, 2016.

\bibitem[\protect\citeauthoryear{He \bgroup \em et al.\egroup
  }{2018}]{he2018soft}
Yang He, Guoliang Kang, Xuanyi Dong, Yanwei Fu, and Yi~Yang.
\newblock Soft filter pruning for accelerating deep convolutional neural
  networks.
\newblock {\em arXiv preprint arXiv:1808.06866}, 2018.

\bibitem[\protect\citeauthoryear{Jiang \bgroup \em et al.\egroup
  }{2021}]{jiang2021all}
Zihang Jiang, Qibin Hou, Li~Yuan, Daquan Zhou, Yujun Shi, Xiaojie Jin, Anran
  Wang, and Jiashi Feng.
\newblock All tokens matter: Token labeling for training better vision
  transformers.
\newblock {\em arXiv preprint arXiv:2104.10858}, 2021.

\bibitem[\protect\citeauthoryear{Kim \bgroup \em et al.\egroup
  }{2021}]{kim2021learned}
Sehoon Kim, Sheng Shen, David Thorsley, Amir Gholami, Woosuk Kwon, Joseph
  Hassoun, and Kurt Keutzer.
\newblock Learned token pruning for transformers.
\newblock {\em arXiv preprint arXiv:2107.00910}, 2021.

\bibitem[\protect\citeauthoryear{Li \bgroup \em et al.\egroup
  }{2016}]{li2016pruning}
Hao Li, Asim Kadav, Igor Durdanovic, Hanan Samet, and Hans~Peter Graf.
\newblock Pruning filters for efficient convnets.
\newblock {\em arXiv preprint arXiv:1608.08710}, 2016.

\bibitem[\protect\citeauthoryear{Liang \bgroup \em et al.\egroup
  }{2022}]{liang2022evit}
Youwei Liang, Chongjian GE, Zhan Tong, Yibing Song, Jue Wang, and Pengtao Xie.
\newblock {E}vit: Expediting vision transformers via token reorganizations.
\newblock In {\em International Conference on Learning Representations (ICLR)},
  2022.

\bibitem[\protect\citeauthoryear{Liu \bgroup \em et al.\egroup
  }{2017}]{Liu2017learning}
Zhuang Liu, Jianguo Li, Zhiqiang Shen, Gao Huang, Shoumeng Yan, and Changshui
  Zhang.
\newblock Learning efficient convolutional networks through network slimming.
\newblock In {\em ICCV}, 2017.

\bibitem[\protect\citeauthoryear{Liu \bgroup \em et al.\egroup
  }{2021}]{liu2021swin}
Ze~Liu, Yutong Lin, Yue Cao, Han Hu, Yixuan Wei, Zheng Zhang, Stephen Lin, and
  Baining Guo.
\newblock Swin transformer: Hierarchical vision transformer using shifted
  windows.
\newblock In {\em Proceedings of the IEEE/CVF International Conference on
  Computer Vision (ICCV)}, pages 10012--10022, 2021.

\bibitem[\protect\citeauthoryear{Liu \bgroup \em et al.\egroup
  }{2022a}]{liu2022dynamic}
Kai Liu, Tianyi Wu, Cong Liu, and Guodong Guo.
\newblock Dynamic group transformer: A general vision transformer backbone with
  dynamic group attention.
\newblock {\em arXiv preprint arXiv:2203.03937}, 2022.

\bibitem[\protect\citeauthoryear{Liu \bgroup \em et al.\egroup
  }{2022b}]{liu2022convnet}
Zhuang Liu, Hanzi Mao, Chao-Yuan Wu, Christoph Feichtenhofer, Trevor Darrell,
  and Saining Xie.
\newblock A convnet for the 2020s.
\newblock {\em Proceedings of the IEEE/CVF Conference on Computer Vision and
  Pattern Recognition (CVPR)}, 2022.

\bibitem[\protect\citeauthoryear{Luo \bgroup \em et al.\egroup
  }{2017}]{luo2017thinet}
Jian-Hao Luo, Jianxin Wu, and Weiyao Lin.
\newblock Thinet: A filter level pruning method for deep neural network
  compression.
\newblock In {\em Proceedings of the IEEE international conference on computer
  vision}, pages 5058--5066, 2017.

\bibitem[\protect\citeauthoryear{Michel \bgroup \em et al.\egroup
  }{2019}]{michel2019sixteen}
Paul Michel, Omer Levy, and Graham Neubig.
\newblock Are sixteen heads really better than one?
\newblock {\em Advances in neural information processing systems}, 32, 2019.

\bibitem[\protect\citeauthoryear{Molchanov \bgroup \em et al.\egroup
  }{2016}]{molchanov2016pruning}
Pavlo Molchanov, Stephen Tyree, Tero Karras, Timo Aila, and Jan Kautz.
\newblock Pruning convolutional neural networks for resource efficient
  inference.
\newblock {\em arXiv preprint arXiv:1611.06440}, 2016.

\bibitem[\protect\citeauthoryear{Pan \bgroup \em et al.\egroup
  }{2021}]{pan2021ia}
Bowen Pan, Rameswar Panda, Yifan Jiang, Zhangyang Wang, Rogerio Feris, and Aude
  Oliva.
\newblock Ia-red$^{2}$: Interpretability-aware redundancy reduction for vision
  transformers.
\newblock In {\em Advances in Neural Information Processing Systems (NeurIPS)},
  2021.

\bibitem[\protect\citeauthoryear{Rao \bgroup \em et al.\egroup
  }{2021}]{rao2021dynamicvit}
Yongming Rao, Wenliang Zhao, Benlin Liu, Jiwen Lu, Jie Zhou, and Cho-Jui Hsieh.
\newblock Dynamicvit: Efficient vision transformers with dynamic token
  sparsification.
\newblock In {\em Advances in Neural Information Processing Systems (NeurIPS)},
  2021.

\bibitem[\protect\citeauthoryear{Tan and Le}{2019}]{tan2019efficientnet}
Mingxing Tan and Quoc Le.
\newblock Efficientnet: Rethinking model scaling for convolutional neural
  networks.
\newblock In {\em ICML}, pages 6105--6114. PMLR, 2019.

\bibitem[\protect\citeauthoryear{Tang \bgroup \em et al.\egroup
  }{2021}]{tang2021patch}
Yehui Tang, Kai Han, Yunhe Wang, Chang Xu, Jianyuan Guo, Chao Xu, and Dacheng
  Tao.
\newblock Patch slimming for efficient vision transformers.
\newblock {\em arXiv preprint arXiv:2106.02852}, 2021.

\bibitem[\protect\citeauthoryear{Touvron \bgroup \em et al.\egroup
  }{2021a}]{touvron2021training}
Hugo Touvron, Matthieu Cord, Matthijs Douze, Francisco Massa, Alexandre
  Sablayrolles, and Herv{\'e} J{\'e}gou.
\newblock Training data-efficient image transformers \& distillation through
  attention.
\newblock In {\em International Conference on Machine Learning (ICML)}, pages
  10347--10357, 2021.

\bibitem[\protect\citeauthoryear{Touvron \bgroup \em et al.\egroup
  }{2021b}]{touvron2021cait}
Hugo Touvron, Matthieu Cord, Alexandre Sablayrolles, Gabriel Synnaeve, and
  Herv{\'e} J{\'e}gou.
\newblock Going deeper with image transformers.
\newblock {\em arXiv preprint arXiv:2103.17239}, 2021.

\bibitem[\protect\citeauthoryear{Vaswani \bgroup \em et al.\egroup
  }{2017}]{transformer}
Ashish Vaswani, Noam Shazeer, Niki Parmar, Jakob Uszkoreit, Llion Jones,
  Aidan~N Gomez, Lukasz Kaiser, and Illia Polosukhin.
\newblock Attention is all you need.
\newblock In {\em NeurIPS}, pages 5998--6008, 2017.

\bibitem[\protect\citeauthoryear{Wang \bgroup \em et al.\egroup
  }{2021}]{wang2021pyramid}
Wenhai Wang, Enze Xie, Xiang Li, Deng-Ping Fan, Kaitao Song, Ding Liang, Tong
  Lu, Ping Luo, and Ling Shao.
\newblock Pyramid vision transformer: A versatile backbone for dense prediction
  without convolutions.
\newblock In {\em Proceedings of the IEEE/CVF International Conference on
  Computer Vision (ICCV)}, pages 568--578, 2021.

\bibitem[\protect\citeauthoryear{Wu \bgroup \em et al.\egroup
  }{2021}]{wu2021cvt}
Haiping Wu, Bin Xiao, Noel Codella, Mengchen Liu, Xiyang Dai, Lu~Yuan, and Lei
  Zhang.
\newblock Cvt: Introducing convolutions to vision transformers.
\newblock In {\em Proceedings of the IEEE/CVF International Conference on
  Computer Vision}, pages 22--31, 2021.

\bibitem[\protect\citeauthoryear{Xie \bgroup \em et al.\egroup
  }{2021}]{xie2021segformer}
Enze Xie, Wenhai Wang, Zhiding Yu, Anima Anandkumar, Jose~M Alvarez, and Ping
  Luo.
\newblock Segformer: Simple and efficient design for semantic segmentation with
  transformers.
\newblock {\em arXiv preprint arXiv:2105.15203}, 2021.

\bibitem[\protect\citeauthoryear{Xu \bgroup \em et al.\egroup
  }{2021}]{xu2021co}
Weijian Xu, Yifan Xu, Tyler Chang, and Zhuowen Tu.
\newblock Co-scale conv-attentional image transformers.
\newblock In {\em Proceedings of the IEEE/CVF International Conference on
  Computer Vision}, pages 9981--9990, 2021.

\bibitem[\protect\citeauthoryear{Xu \bgroup \em et al.\egroup
  }{2022}]{xu2022evovit}
Yifan Xu, Zhijie Zhang, Mengdan Zhang, Kekai Sheng, Ke~Li, Weiming Dong, Liqing
  Zhang, Changsheng Xu, and Xing Sun.
\newblock Evo-vit: Slow-fast token evolution for dynamic vision transformer.
\newblock In {\em Proceedings of the AAAI Conference on Artificial Intelligence
  (AAAI)}, 2022.

\bibitem[\protect\citeauthoryear{Yang \bgroup \em et al.\egroup
  }{2021}]{yang2021nvit}
Huanrui Yang, Hongxu Yin, Pavlo Molchanov, Hai Li, and Jan Kautz.
\newblock Nvit: Vision transformer compression and parameter redistribution.
\newblock {\em arXiv preprint arXiv:2110.04869}, 2021.

\bibitem[\protect\citeauthoryear{Yin \bgroup \em et al.\egroup
  }{2021}]{yin2021adavit}
Hongxu Yin, Arash Vahdat, Jose Alvarez, Arun Mallya, Jan Kautz, and Pavlo
  Molchanov.
\newblock Adavit: Adaptive tokens for efficient vision transformer.
\newblock {\em arXiv preprint arXiv:2112.07658}, 2021.

\bibitem[\protect\citeauthoryear{Yu \bgroup \em et al.\egroup
  }{2022}]{yu2022unified}
Shixing Yu, Tianlong Chen, Jiayi Shen, Huan Yuan, Jianchao Tan, Sen Yang,
  Ji~Liu, and Zhangyang Wang.
\newblock Unified visual transformer compression.
\newblock {\em arXiv preprint arXiv:2203.08243}, 2022.

\bibitem[\protect\citeauthoryear{Yuan \bgroup \em et al.\egroup
  }{2021}]{Yuan_2021_ICCV}
Li~Yuan, Yunpeng Chen, Tao Wang, Weihao Yu, Yujun Shi, Zi-Hang Jiang,
  Francis~E.H. Tay, Jiashi Feng, and Shuicheng Yan.
\newblock Tokens-to-token vit: Training vision transformers from scratch on
  imagenet.
\newblock In {\em Proceedings of the IEEE/CVF International Conference on
  Computer Vision (ICCV)}, pages 558--567, October 2021.

\bibitem[\protect\citeauthoryear{Zheng \bgroup \em et al.\egroup }{2021}]{SETR}
Sixiao Zheng, Jiachen Lu, Hengshuang Zhao, Xiatian Zhu, Zekun Luo, Yabiao Wang,
  Yanwei Fu, Jianfeng Feng, Tao Xiang, Philip~H.S. Torr, and Li~Zhang.
\newblock Rethinking semantic segmentation from a sequence-to-sequence
  perspective with transformers.
\newblock In {\em CVPR}, 2021.

\bibitem[\protect\citeauthoryear{Zhu \bgroup \em et al.\egroup
  }{2020}]{zhu2020deformable}
Xizhou Zhu, Weijie Su, Lewei Lu, Bin Li, Xiaogang Wang, and Jifeng Dai.
\newblock Deformable detr: Deformable transformers for end-to-end object
  detection.
\newblock {\em arXiv preprint arXiv:2010.04159}, 2020.

\bibitem[\protect\citeauthoryear{Zhu \bgroup \em et al.\egroup
  }{2021}]{zhu2021vision}
Mingjian Zhu, Yehui Tang, and Kai Han.
\newblock Vision transformer pruning.
\newblock {\em arXiv preprint arXiv:2104.08500}, 2021.

\end{thebibliography}

\clearpage
\appendix
\section{Experiment details}
\label{sec:detail}
We adopt the same fine-tuning strategy as EViT~\cite{liang2022evit}, and set the initial learning rate to 2e-5 and decrease to 2e-6 as cosine, while the weight decay is set to 1e-6. For DeiT \cite{touvron2021training}, we insert learnable thresholds after the 4th, 7th and 10th layers. For LV-ViT \cite{jiang2021all}, we insert thresholds after the 5th, 9th, and 13th layers. We empirically set the temperature parameter T of the sigmoid function used to generate the soft binary pruning mask to 1e+4. For all models, the initialized values of learnable thresholds are set to [0.001, 0.002, 0.003].


\section{More Analysis}
\paragraph{Comparisons with other models.} We compare the complexity and accuracy of our adaptive sparse LV-ViT (abbreviated as AS-LV-ViT) with other sota models on ImageNet in Table \ref{tab:sota}.

\paragraph{Comparison with random baseline.} We compare our method with a random and minimal baseline based on the pruning ratios obtained from adaptive pruning. The pruning ratio was first obtained by comparing the trained threshold with the token scores, and then two scenarios were experimented with randomly sampled tokens (random) and the lowest scoring k tokens (minK). The results in the Table \ref{tab:rand} demonstrate the effectiveness of our method. The minK method discards a large number of information tokens, causing significant accuracy degradation, while random baselines perform moderately well at low compression rates and show obvious accuracy drop when the compression rate is increased. In contrast, our method retains the most informative tokens and achieves a precision-speed balance.

\begin{table}[htbp]
\centering
\caption{Results of the random baseline.}
\begin{tabular}{ccc}
    \toprule
    Method     & GFLOPs & Acc (\%)  \\ \midrule
    minK    & 3.0    & 51.63    \\ \midrule
    random      & 3.0    & 72.79    \\ \midrule
    ours    & 3.0    & 79.63   \\ \midrule \midrule
    minK    & 2.3    & 26.65    \\ \midrule
    random      & 2.3    & 62.67    \\ \midrule
    ours    & 2.3    & 78.70   \\
    \bottomrule
\end{tabular}
\label{tab:rand}
\end{table}

\paragraph{Effect of random seeds.} We tried four random seeds in the DeiT-S model and the experimental results are shown in the Table \ref{tab:seed}, indicating that our method is stable.

\begin{table}[htbp]
\centering
\caption{Accuracy under different random seeds.}
\begin{tabular}{ccccc}
    \toprule
    seed     & s1 & s2 & s3 & s4 \\ \midrule
    Acc (\%)    & 79.63    & 79.60     & 79.57    & 79.62   \\
    \bottomrule
\end{tabular}
\label{tab:seed}
\end{table}

\begin{algorithm}[htbp]
\caption{Adaptive Sparsity Module in a PyTorch-like style.}
\label{alg:code}
\definecolor{codeblue}{rgb}{0.25,0.5,0.5}
\lstset{
  backgroundcolor=\color{white},
  basicstyle=\fontsize{8.8pt}{8.8pt}\ttfamily\selectfont,
  columns=fullflexible,
  breaklines=true,
  captionpos=b,
  commentstyle=\fontsize{7.2pt}{7.2pt}\color{codeblue},
  keywordstyle=\fontsize{7.2pt}{7.2pt}
}
\begin{lstlisting}[language=python]
# score: token score
# thres: learnable threshold
# tau: temperature parameter
# training: training/inference mode 

def asm(score, thres, tau=1, training=False):
    # batch size
    B = score.size(0)
    # inference mode
    if not training:
        idx = logits > thres
        if B==1:
            return idx
        k = idx.sum().item() // B
        return k
    
    # training mode
    y_soft = torch.sigmoid((score - thres) * tau)
    y_hard = (y_soft > 0.5).float()
    # straight through
    mask = y_hard - y_soft.detach() + y_soft

    return mask

\end{lstlisting}
\end{algorithm}

\section{Visualization}
\begin{figure*}[htbp]
\centering
\includegraphics[width=1\textwidth]{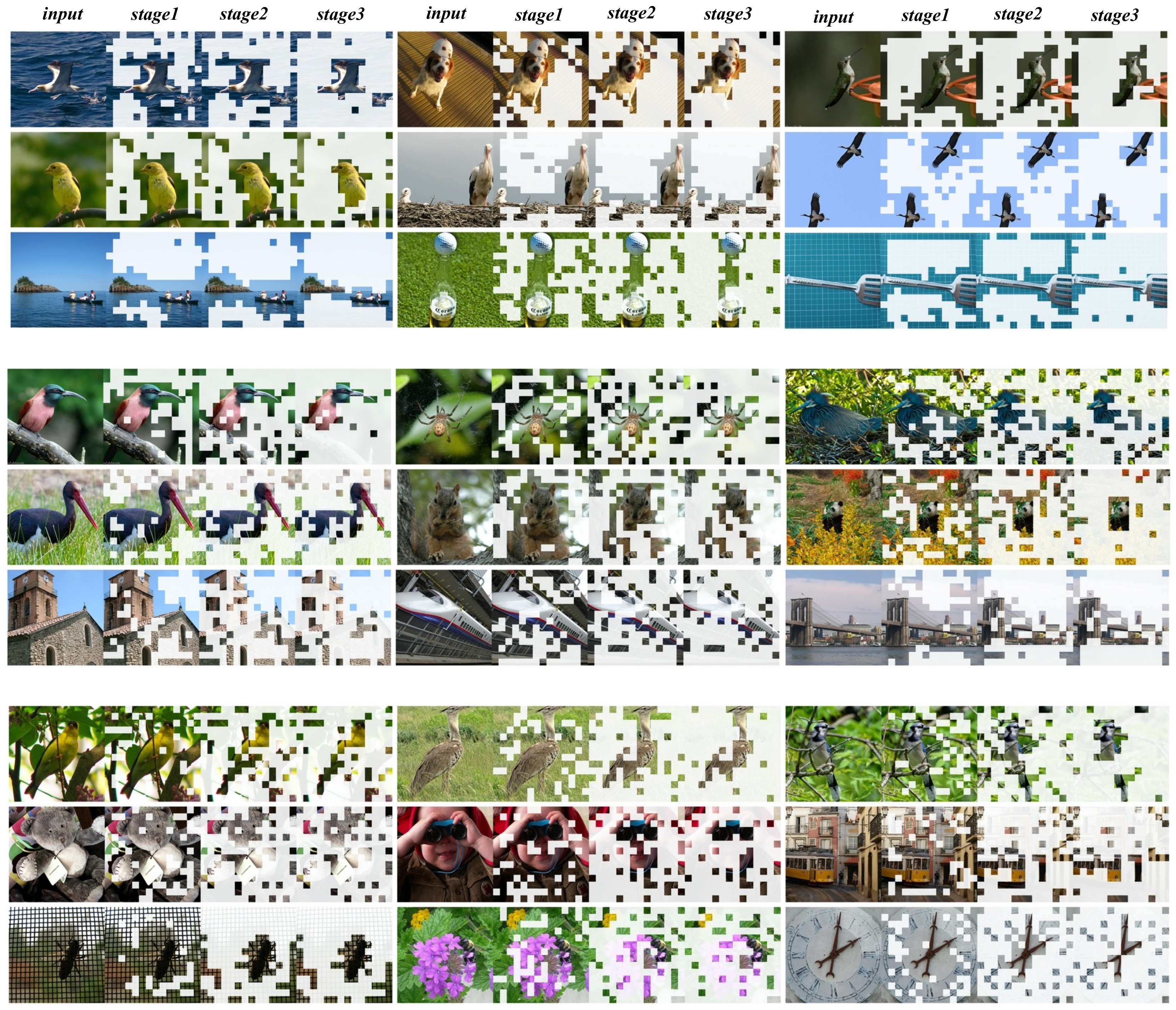}
\caption{Effects of token pruning for different difficulty images at three stages. Easy, normal and complicated samples are shown in order from top to bottom, separated by blank lines. Easy samples are discarded with lots of uninformative tokens in the first stage by our model, and the pruning rate of common samples gradually increases in each stage. Complex samples are pruned with only limited number of tokens in the initial stage, and useless tokens are discarded in the later stages as the confidence of the model increases.}
\label{fig:imgviz}
\end{figure*}

We obtained all pruning configurations for the AS-DeiT-S/$f=$0.65 model on the ImageNet validation dataset and ranked them by the first stage's token number. Then select the images that are easy, normal and complicated for our model and visualize their pruning effects. As shown in the Figure \ref{fig:imgviz}, our method prunes large numbers of uninformative tokens early for simple samples to save computational effort, uses the regular strategy for ordinary samples, and gradually increases the pruning rate for complex samples to ensure the recognition accuracy.

\begin{table*}[htbp]
  \centering
  \caption{Comparisons of model complexity and accuracy on ImageNet with advanced CNN and ViT models. Throughput and latency are measured on the same machine.}
    \begin{tabu}to\textwidth{l*{5}{X[c]}}\toprule
    Model & Params (M) & GFLOPs & Top-1 Acc (\%) & Throughput (img/s) & Latency (ms) \\\midrule
    DeiT-S~\cite{touvron2021training} & 22.1  & 4.6  & 79.8  & 770  & 6.16  \\
    PVT-S~\cite{wang2021pyramid} & 24.5  & 3.8  & 79.8  & 558  & 12.10 \\
    Twins-PCPVT-S~\cite{chu2021twins} & 24.1  & 3.8  & 81.2  & 557  & 11.54 \\
    CoaT-Lite Small~\cite{xu2021co} & 19.8  & 4.0  & 81.9  & 368  & 15.68 \\
    Swin-T~\cite{liu2021swin} & 29.0  & 4.5  & 81.3  & 533  & 8.83 \\
    CvT-13~\cite{wu2021cvt} & 20.0  & 4.5  & 81.6 & 527  & 15.0 \\
    ConvNeXt-T~\cite{liu2022convnet} & 28.6  & 4.5  & 82.1 & 332  & 5.39 \\
    T2T-ViT-14~\cite{Yuan_2021_ICCV} & 22.0  & 5.2  & 81.5 & 601  & 8.78 \\
    \textbf{AS-LV-S/0.6} (Ours) & 26.2  & 3.9  & \textbf{82.6}  & \textbf{1023}  & 9.31 \\
    \midrule \midrule
    PVT-M~\cite{wang2021pyramid} & 44.2  & 6.7  & 81.2  & 362  & 19.49 \\
    Twins-PCPVT-B~\cite{chu2021twins} & 43.8  & 6.7  & 82.7  & 347  & 19.9 \\
    CvT-21~\cite{wu2021cvt} & 32.0  & 7.1  & 82.5 & 340  & 25.10 \\
    Twins-SVT-B~\cite{chu2021twins} & 56.1  & 8.6  & 83.2  & 313  & 16.15 \\
    ConvNeXt-S~\cite{liu2022convnet} & 50.2  & 8.7  & 83.1 & 192  & 9.94 \\
    Swin-S~\cite{liu2021swin} & 49.6  & 8.7  & 83.0  & 308  & 17.0 \\
    T2T-ViT-19~\cite{Yuan_2021_ICCV} & 39.2  & 8.9  & 81.9 & 375  & 11.66 \\
    CaiT-S-24~\cite{touvron2021cait} & 46.9  & 9.4  & 82.7 & 245  & 18.68 \\
    PVT-L~\cite{wang2021pyramid} & 61.4  & 9.8  & 81.7  & 249  & 28.3 \\
    CoaT-Lite Medium~\cite{xu2021co} & 19.8  & 9.6  & 83.6  & 368  & 15.68 \\
    EfficientNet-B5$^*$/456~\cite{tan2019efficientnet} & 30.0  & 9.9  & 83.6  & 97  & 17.51 \\
    \textbf{AS-LV-M/0.67} (Ours) & 55.8  & 8.5  & \textbf{83.9}  & \textbf{801}  & 11.53
    \\\bottomrule
    \end{tabu}%
  \label{tab:sota}%
\end{table*}%

\end{document}